\documentclass[journal]{IEEEtran}

\ifCLASSINFOpdf
\else
   \usepackage[dvips]{graphicx}
\fi
\usepackage{url}
\usepackage{graphicx}
\usepackage{xcolor}
\usepackage{hyperref}
\usepackage{amsmath}
\usepackage{amssymb}
\usepackage{booktabs}
\usepackage{adjustbox}
\usepackage{colortbl}
\usepackage{multirow}

\begin{document}

\title{Multispectral Detection Transformer with Infrared-Centric Feature Fusion}

\author{Seongmin Hwang, Daeyoung Han,  
        and Moongu Jeon, \IEEEmembership{Senior Member, IEEE}
\thanks{
    S. Hwang is with the Artificial Intelligence Graduate School, Gwangju Institute of Science and Technology (GIST), Gwangju 61005, Republic of Korea (e-mail: sm.hwang@gm.gist.ac.kr).

    D. Han, and M. Jeon are with the School of Electrical Engineering and Computer Science, Gwangju Institute of Science and Technology (GIST), Gwangju 61005, Republic of Korea (e-mail: xesta120@gist.ac.kr; mgjeon@gist.ac.kr).

    (Corresponding author: Moongu Jeon).
    }}

\markboth{Journal of \LaTeX\ Class Files, Vol. 14, No. 8, August 2015}
{Shell \MakeLowercase{\textit{et al.}}: Bare Demo of IEEEtran.cls for IEEE Journals}
\maketitle
\begin{abstract}
Multispectral object detection aims to leverage complementary information from visible (RGB) and infrared (IR) modalities to enable robust performance under diverse environmental conditions. Our key insight, derived from wavelet analysis and empirical observations, is that IR images contain structurally rich high-frequency information critical for object detection, making an infrared-centric approach highly effective. To capitalize on this finding, we propose Infrared-Centric Fusion (IC-Fusion), a lightweight and modality-aware sensor fusion method that prioritizes infrared features while effectively integrating complementary RGB semantic context. IC-Fusion adopts a compact RGB backbone and designs a novel fusion module comprising a Multi-Scale Feature Distillation (MSFD) block to enhance RGB features and a three-stage fusion block with a Cross-Modal Channel Shuffle Gate (CCSG), a Cross-Modal Large Kernel Gate (CLKG), and a Channel Shuffle Projection (CSP) to facilitate effective cross-modal interaction. Experiments on the FLIR and LLVIP benchmarks demonstrate the superior effectiveness and efficiency of our IR-centric fusion strategy, further validating its benefits. Our code is available at \href{https://github.com/smin-hwang/IC-Fusion}{https://github.com/smin-hwang/IC-Fusion}.
\end{abstract}


\IEEEpeerreviewmaketitle

\section{Introduction}
\label{sec:intro}

\IEEEPARstart{O}{bject} detection in adverse visual environments, such as night-time, foggy, and low-light conditions, remains a significant challenge in computer vision. Although the visible (RGB) sensor provides rich texture and color information under favorable lighting, its performance often degrades significantly in poorly illuminated scenes. In contrast, infrared (IR) imagery captures thermal radiation emitted by objects, offering complementary and often more robust information that is less affected by ambient lighting variations. The fusion of RGB and IR modalities has thus become a promising direction to enhance object detection under diverse conditions.

\begin{figure}[t]
 \centering
    \includegraphics[scale=0.18]{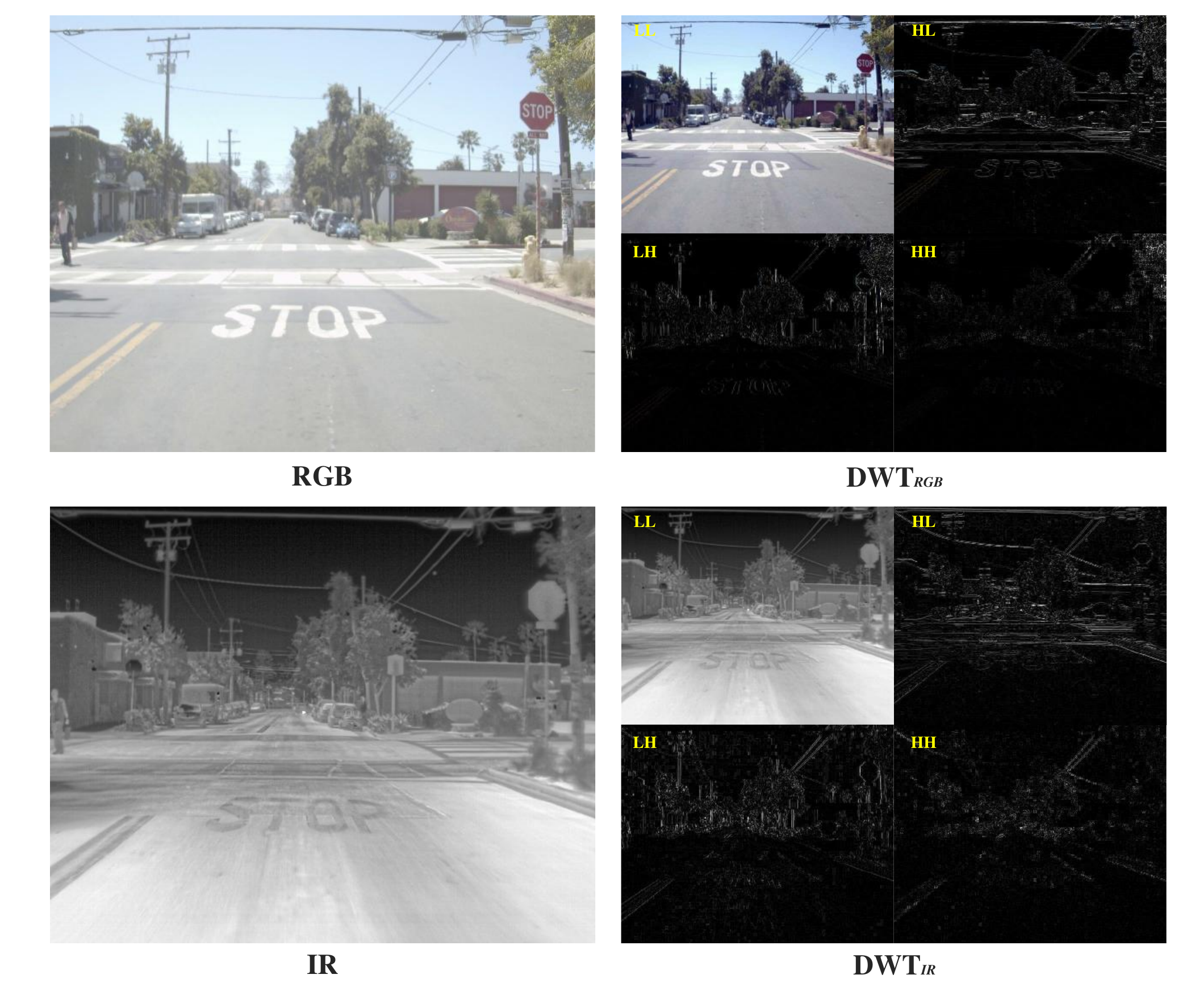}
    \caption{Wavelet decomposition of visible (RGB) and infrared (IR) images.}
    \label{fig:wavelet_analysis}
    \vspace{-0.5cm}
 \end{figure}

Despite the potential of multispectral object detection~\cite{zhang2021guided, li2023stabilizing, zhu2023multi, xing2023multispectral, xiao2024gm}, effectively integrating the two modalities remains challenging due to their inherent spectral differences. To better understand the contribution of each modality, we perform a wavelet decomposition on RGB and IR images, as shown in Fig.~\ref{fig:wavelet_analysis}. The RGB image primarily contains low-frequency semantic structures in the LL band. In contrast, the IR modality presents distinct object boundaries and contours, especially in the LH and HL sub-bands, which are crucial for robust localization~\cite{gidaris2016locnet} and shape understanding~\cite{zhu2015segdeepm}. This observation highlights that IR images inherently encode sharper structural cues that are highly beneficial for object detection, particularly in edge and boundary regions.

The effectiveness of IR features is further quantitatively validated in Table~\ref{tab:rgb_ir_comparison}, where unimodal IR detectors consistently outperform their RGB counterparts on the FLIR-aligned~\cite{zhang2020multispectral} benchmark across various backbones. For example, RT-DETR~\cite{zhao2024detrs} trained solely on IR data achieves a mean Average Precision (mAP) of 43.6, significantly surpassing its RGB equivalent (33.5 mAP). These findings imply that heavy feature extraction from the RGB modality may be redundant and detection performance can be retained or improved, even with significantly reduced computation by relying on IR-centric representations and adopting a lightweight RGB backbone.

Motivated by these insights, we propose a multispectral object detection transformer, IC-Fusion, which incorporates a novel cross-modal fusion module to efficiently fuse information across modalities. To reduce redundancy and improve efficiency, we adopt a lightweight backbone for the RGB modality while preserving a deeper IR representation to retain structural details. Our design emphasizes the dominant structural contributions of the IR modality while enhancing the complementary semantic information from RGB. Specifically, we introduce a Multi-Scale Feature Distillation (MSFD) module to enrich RGB features with multi-scale spatial context, and adopt a three-stage fusion block consisting of a Cross-Modal Channel Shuffle Gate (CCSG), a Cross-Modal Large Kernel Gate (CLKG), and a Channel Shuffle Projection (CSP) module to effectively extract and fuse complementary cross-modal representations. Experiments on benchmark datasets validate the effectiveness of our design and further highlight the advantage of IR-centric fusion for efficient multispectral object detection.

 \begin{table}[ht]
\centering
\caption{Comparisons on the FLIR-aligned dataset. We report mAP@50, mAP@75, and mAP under IoU thresholds.}
\renewcommand{\arraystretch}{0.9}
\begin{tabular}{l c c c c}
\toprule
\textbf{Model} & \textbf{Modality} & \textbf{mAP50} & \textbf{mAP75} & \textbf{mAP} \\
\midrule
Faster-RCNN~\cite{ren2015faster} & RGB & 64.9 & 21.1 & 28.9 \\
Yolov5~\cite{qingyun2021cross} & RGB & 67.8 & 25.9 & 31.8 \\
RT-DETR~\cite{zhao2024detrs} & RGB & 69.7 & 26.5 & 33.5 \\
\midrule
Faster-RCNN~\cite{ren2015faster} & IR & 74.4 & 32.5 & 37.6 \\
Yolov5~\cite{qingyun2021cross} & IR & 80.1 & - & 42.4 \\
RT-DETR~\cite{zhao2024detrs} & IR & 80.5 & 39.6 & 43.6 \\
\bottomrule
\end{tabular}
\label{tab:rgb_ir_comparison}
\end{table}

 \begin{figure}[ht]
 \centering
    \includegraphics[scale=0.20]{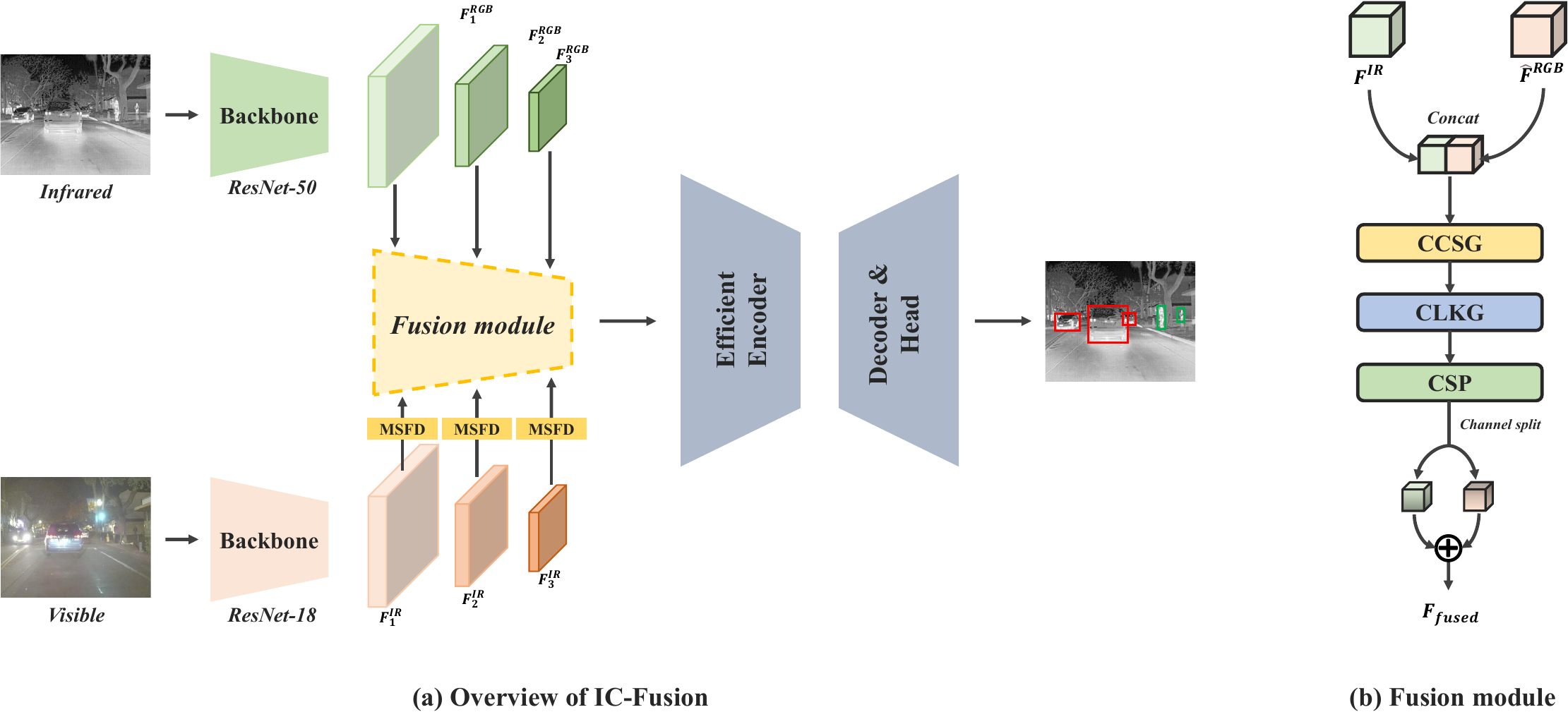}
    \caption{Overview of (a) the proposed IC-Fusion framework and (b) the proposed cross-modal feature fusion module composed of CCSG, CLKG, and CSP modules.}
    \label{fig:architecture}
        \vspace{-0.5cm}
 \end{figure}

\section{Proposed Method}
\label{sec:method}

\subsection{Overview}

We propose IC-Fusion, a multispectral object detection transformer framework that effectively integrates RGB and IR features through structured cross-modal fusion. As shown in Fig.~\ref{fig:architecture}, the framework consists of three main components: dual-stream backbones, a fusion module, and a DETR-based transformer encoder-decoder~\cite{zhao2024detrs, DETR, zhudeformable, zhangdino}.

Given a pair of RGB and IR images \( (x_{\text{rgb}}, x_{\text{ir}}) \), we extract features using two modality-specific CNN backbones. A lightweight ResNet-18 is used for the RGB modality to ensure efficiency, while a ResNet-50 is adopted for the IR modality due to its robustness under adverse factors such as low illumination, occlusions, and weather conditions and its ability to capture thermal contours and object shapes.

From each backbone, we extract multi-scale features from the last three stages:
\begin{equation}
\mathbb{F}_{\text{rgb}} = \left\{ \mathbf{F}_{\text{rgb}}^{(3)}, \mathbf{F}_{\text{rgb}}^{(4)}, \mathbf{F}_{\text{rgb}}^{(5)} \right\}, \quad
\mathbb{F}_{\text{ir}} = \left\{ \mathbf{F}_{\text{ir}}^{(3)}, \mathbf{F}_{\text{ir}}^{(4)}, \mathbf{F}_{\text{ir}}^{(5)} \right\}
\end{equation}

To enhance the representational quality and ensure channel alignment between modalities, each RGB feature \( \mathbf{F}_{\text{rgb}}^{(l)} \) is first processed through a lightweight refinement module:
\begin{equation}
\hat{\mathbf{F}}_{\text{rgb}}^{(l)} = \mathcal{R} \left( \mathbf{F}_{\text{rgb}}^{(l)} \right), \quad l \in \{3, 4, 5\}
\end{equation}
where \( \mathcal{R}(\cdot) \) denotes the refinement operation implemented by the Multi-Scale Feature Distillation (MSFD) module

Each pair of refined RGB and IR features \( \left( \hat{\mathbf{F}}_{\text{rgb}}^{(l)}, \mathbf{F}_{\text{ir}}^{(l)} \right) \) is then fused via a cross-modal fusion block:
\begin{equation}
\mathbf{F}_{\text{fused}}^{(l)} = \mathcal{F} \left( \hat{\mathbf{F}}_{\text{rgb}}^{(l)}, \mathbf{F}_{\text{ir}}^{(l)} \right)
\end{equation}
where \( \mathcal{F}(\cdot) \) represents the cross-modal fusion process, which sequentially applies a Cross-Modal Channel Shuffle Gate (CCSG) for feature interaction, a Cross-Modal Large Kernel Gate (CLKG) for long-range contextual alignment, and a Channel Shuffle Projection (CSP) module. The output of this three-stage fusion block then undergoes a channel split into two parallel feature streams, which are subsequently element-wise added to form the final fused feature map. This design facilitates the integration of cross-modal information, thereby enhancing feature richness and training stability by enabling the model to selectively propagate refined features. 

The fused output integrates complementary semantics from both modalities while suppressing redundant information, and is subsequently passed to an efficient encoder~\cite{zhao2024detrs}, followed by a DETR-decoder for final object prediction. This design enables efficient and robust multispectral detection under challenging conditions.

 \begin{figure}[ht]
 \centering
    \includegraphics[scale=0.25]{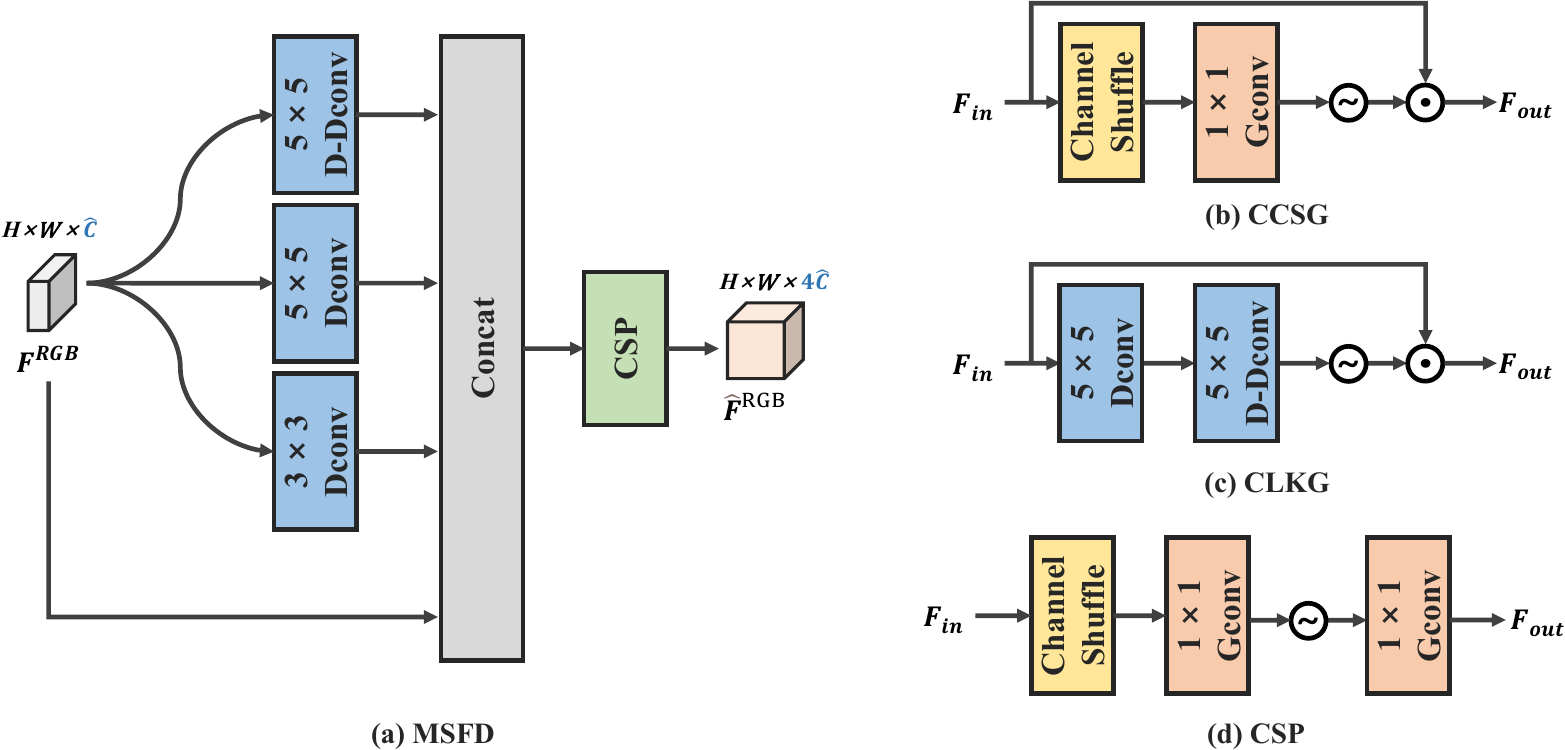}
    \caption{Illustration of key components in the proposed framework. (a) Multi-Scale Feature Distillation (MSFD) module that enhances RGB features using multi-branch depthwise convolutions, channel shuffle, and grouped projection. (b) Cross-Modal Channel Shuffle Gate (CCSG) for gated channel-wise interaction. (c) Cross-Modal Large Kernel Gate (CLKG) for capturing long-range structural features. (d) Channel Shuffle Projection (CSP) module for efficient feature projection and distillation.}
    \label{fig:components}
        \vspace{-0.5cm}
 \end{figure}

\subsection{Multi-Scale Feature Distillation}

To enhance the representational capacity of the RGB features while aligning their dimensions with the IR stream, we introduce the Multi-Scale Feature Distillation (MSFD) module. As illustrated in Fig.~\ref{fig:components}(a), the MSFD module is designed to extract spatially diverse cues from multiple receptive fields in an efficient and compact manner.

Given an input RGB feature map \( \mathbf{F}_{\text{RGB}} \in \mathbb{R}^{H \times W \times \hat{C}} \), we apply a multi-branch depthwise convolution block \( \mathcal{W}(\cdot) \) that captures local and contextual information using kernel sizes \(3 \times 3\), \(5 \times 5\), and dilated \(5 \times 5\) with dilation rate 2. This operation produces a set of multi-scale features:
\begin{equation}
\left\{ \mathbf{F}_{k} \right\}_{k \in \{3\times3,\,5\times5,\,5\times5^{\text{dil}} \}} = \mathcal{W}(\mathbf{F}_{\text{RGB}})
\end{equation}

The resulting features are concatenated with the original input along the channel dimension:
\begin{equation}
\mathbf{F}_{\text{cat}} = \text{Concat} \left( \mathbf{F}_{\text{RGB}}, \mathbf{F}_{3\times3}, \mathbf{F}_{5\times5}, \mathbf{F}_{5\times5^{\text{dil}}} \right) \in \mathbb{R}^{H \times W \times 4\hat{C}}
\end{equation}

\begin{table*}[ht]
\centering
\caption{Performance comparison on FLIR-aligned and LLVIP datasets. We report mAP@50, mAP@75, and overall mAP. Our method achieves the best performance on LLVIP dataset.}
\renewcommand{\arraystretch}{0.9}
\begin{tabular}{l c l l l | c c c | c c c}
\toprule
\multirow{2}{*}{\textbf{Model}} & \multirow{2}{*}{\textbf{Venue \& Year}} & \multirow{2}{*}{\textbf{RGB Backbone}} & \multirow{2}{*}{\textbf{IR Backbone}} & \multirow{2}{*}{\textbf{Modality}} &
\multicolumn{3}{c|}{\textbf{FLIR-aligned}} & 
\multicolumn{3}{c}{\textbf{LLVIP}} \\
 & & & & & mAP50 & mAP75 & mAP & mAP50 & mAP75 & mAP \\
\midrule
Yolov5~\cite{qingyun2021cross} & - & CSPDarknet53 & - & RGB & 67.8 & 25.9 & 31.8 & 90.8 & 56.4 & 52.7 \\
Yolov5~\cite{qingyun2021cross} & - & - & CSPDarknet53 & IR & 80.1 & - & 42.4 & 96.5 & 76.4 & 67.0 \\
RT-DETR~\cite{zhao2024detrs} & CVPR'2024 & ResNet50 & - & RGB & 69.7 & 26.5 & 33.5 & 97.3 & 78.4 & 67.9 \\
RT-DETR~\cite{zhao2024detrs} & CVPR'2024 & - & ResNet50 & IR & 80.5 & 39.6 & 43.6 & 91.5 & 59.5 & 54.2 \\
\midrule
CFT~\cite{qingyun2021cross} & - & CSPDarknet53 & CSPDarknet53 & RGB+IR & 78.7 & 35.5 & 40.2 & 97.5 & 72.9 & 63.6 \\
GAFF~\cite{zhang2021guided} & WACV'2021 & ResNet18 & ResNet18 & RGB+IR & 72.9 & 32.9 & 36.6 & - & - & - \\
SMPD~\cite{li2023stabilizing} & TCSVT'2023 & VGG16 & VGG16 & RGB+IR & 73.6 & - & - & - & - & - \\
MFPT~\cite{zhu2023multi} & TITS'2023 & ResNet50 & ResNet50 & RGB+IR & 80.0 & - & - & - & - & - \\
CSSA~\cite{cao2023multimodal} & CVPRW'2023 & ResNet50 & ResNet50 & RGB+IR & 79.2 & 37.4 & 41.3 & 94.3 & 66.6 & 59.2 \\
LRAF-Net~\cite{fu2023lraf} & TNNLS'2023 & CSPDarknet53 & CSPDarknet53 & RGB+IR & 80.5 & - & 42.8 & \textbf{97.9} & - & 66.3 \\
ICAFusion~\cite{shen2024icafusion} & PR'2024 & CSPDarknet53 & CSPDarknet53 & RGB+IR & 79.2 & 36.9 & 41.4 & - & - & - \\
MS-DETR~\cite{xing2023multispectral} & TITS'2024 & ResNet50 & ResNet50 & RGB+IR & - & - & - & \textbf{97.9} & 76.3 & 66.1 \\
GM-DETR~\cite{xiao2024gm} & CVPRW'2024 & ResNet50 & ResNet50 & RGB+IR & \textbf{83.9} & \textbf{42.6} & 45.8 & 97.4 & 81.4 & 70.2 \\
\midrule
\rowcolor{gray!20}
\textbf{IC-Fusion (ours)} & - & ResNet18 & ResNet50 & RGB+IR & 83.3 & \textbf{42.6} & \textbf{46.1} & \textbf{97.9} & \textbf{81.5} & \textbf{70.3} \\
\bottomrule
\end{tabular}
\label{tab:combined_flir_llvip}
\end{table*}

Finally, the concatenated features \( \mathbf{F}_{\text{cat}} \) are passed to a Channel Shuffle Projection (CSP) module for final feature distillation. As illustrated in Fig.~\ref{fig:components}(d), the CSP module is designed to efficiently fuse features across channels. It first applies a channel shuffle operation to enable effective information flow across different channel groups. Subsequently, it employs two sequential grouped $1 \times 1$ convolutions and GELU~\cite{hendrycks2016gaussian} activation to project and distill the features in a computationally lightweight yet expressive manner. The operation of CSP can be formalized as:
\begin{equation}
\hat{\mathbf{F}}_{\text{RGB}} = \mathcal{G}_2 \left( \sigma \left( \mathcal{G}_1 \left( \mathcal{S}(\mathbf{F}_{\text{cat}}) \right) \right) \right)
\end{equation}
where \( \mathcal{S}(\cdot) \) denotes the channel shuffle operation, \( \sigma(\cdot) \) is the GELU activation function, and \( \mathcal{G}_1(\cdot) \) and \( \mathcal{G}_2(\cdot) \) represent the first and second grouped $1 \times 1$ convolutions, respectively. This composite process yields the distilled RGB representation.

This process enables the MSFD module to encode both fine-grained and large-scale context through its multi-branch structure, with the final distillation handled by the CSP module. The enriched RGB representation \( \hat{\mathbf{F}}_{\text{RGB}} \) is then passed to the subsequent fusion stage.

\subsection{Cross-Modal Channel Shuffle Gate}

To facilitate effective feature interaction between RGB and IR modalities, we design the Cross-Modal Channel Shuffle Gate (CCSG) module, illustrated in Fig.~\ref{fig:components}(b). CCSG performs lightweight yet expressive cross-modal refinement through group-based channel reorganization and a gating mechanism~\cite{dauphin2017language}.

Given an input feature map \( \mathbf{F}_{\text{in}} \in \mathbb{R}^{H \times W \times C} \), we first perform channel shuffle to enable cross-modal information exchange:
\begin{equation}
\tilde{\mathbf{F}} = \mathcal{S}(\mathbf{F}_{\text{in}})
\end{equation}

Next, we apply a grouped \(1 \times 1\) convolution to transform the shuffled features:
\begin{equation}
\mathbf{F}_{\text{g}} = \mathcal{G}_{1 \times 1}(\tilde{\mathbf{F}})
\end{equation}

A gating mechanism is then applied by element-wise multiplication between the transformed features and the shuffled input:
\begin{equation}
\mathbf{F}_{\text{gate}} = \sigma(\mathbf{F}_{\text{g}}) \odot \tilde{\mathbf{F}}
\end{equation}
where \( \odot \) denotes element-wise multiplication, and \( \sigma(\cdot) \) is a GELU, which provides smooth and differentiable gating. This allows the model to softly emphasize informative features through cross-modal channel interactions while suppressing redundant and noisy components.

Finally, a residual connection is added to stabilize training:
\begin{equation}
\mathbf{F}_{\text{out}} = \mathbf{F}_{\text{in}} + \mathbf{F}_{\text{gate}}
\end{equation}

The lightweight design of the CCSG module enables efficient use throughout the fusion process without incurring significant overhead.

\subsection{Cross-Modal Large Kernel Gate}

To capture long-range contextual dependencies and enhance structural consistency across modalities, we introduce the Cross-Modal Large Kernel Gate (CLKG) module, as illustrated in Fig.~\ref{fig:components}(c). CLKG is designed to reinforce semantically aligned regions while suppressing modality-specific noise through spatial-gating with large receptive fields.

Given an input fused feature map \( \mathbf{F}_{\text{in}} \in \mathbb{R}^{H \times W \times C} \), we first apply a two-layer large kernel depthwise convolution block, denoted as \( \mathcal{K}(\cdot) \), composed of a standard \(5 \times 5\) and a dilated \(5 \times 5\) depthwise convolution:
\begin{equation}
\mathbf{F}_{\text{ctx}} = \mathcal{K}(\mathbf{F}_{\text{in}})
\end{equation}

The output of the large kernel block is passed through a non-linear activation (GELU) and multiplied element-wise with the original input to form a spatial gate:
\begin{equation}
\mathbf{F}_{\text{gate}} = \sigma(\mathbf{F}_{\text{ctx}}) \odot \mathbf{F}_{\text{in}}
\end{equation}
where \( \sigma(\cdot) \) denotes the GELU activation, and \( \odot \) is element-wise multiplication.

Finally, the gated result is combined with the input via a residual connection:
\begin{equation}
\mathbf{F}_{\text{out}} = \mathbf{F}_{\text{in}} + \mathbf{F}_{\text{gate}}
\end{equation}

This large kernel spatial-gating allows the CLKG module to enhance informative structures and salient regions across modalities. By leveraging a wide receptive field, CLKG effectively complements fine-grained textures from RGB with structural object shapes from IR, thereby enhancing cross-modal representations.

\section{Experiments}

\subsection{Experimental Setup}

We evaluate our method on two widely-used multispectral benchmarks: FLIR~\cite{flir2018adas} and LLVIP~\cite{jia2021llvip}, which provide aligned RGB and IR image pairs captured under diverse illumination conditions, including daytime, night-time, and low-light scenes. The FLIR dataset contains 4,129 training and 1,013 testing image pairs, with annotations for three object classes: \textit{person}, \textit{car}, and \textit{bicycle}. We follow the aligned version curated by~\cite{zhang2020multispectral}. LLVIP includes 12,023 training and 3,462 testing pairs focused on low-light pedestrian detection. Performance is measured using mean Average Precision (mAP) across IoU thresholds from 0.50 to 0.95, and mAP@0.5 for more relaxed localization. 

We adopt RT-DETR~\cite{zhao2024detrs} as the base detector for its efficient design. The ResNet~\cite{he2016deep} backbone is initialized with ImageNet-pretrained weights, while the fusion modules, encoder, and decoder are trained from scratch. Input images are resized to $640 \times 640$, and the model is trained for 60 epochs using the AdamW optimizer. We set the base learning rate to $1 \times 10^{-4}$ and use a smaller learning rate of $1 \times 10^{-5}$ for the backbone to stabilize fine-tuning. Training uses a batch size of 8 and includes basic augmentations such as random horizontal flipping. The overall optimization follows the loss formulation of RT-DETR. To address the limited aligned data, we apply a two-stage training strategy~\cite{xiao2024gm}, where the model is first pretrained on modality-isolated RGB and IR data, followed by fusion training on paired multispectral images.

 \begin{figure}[t]
 \centering
    \includegraphics[scale=0.10]{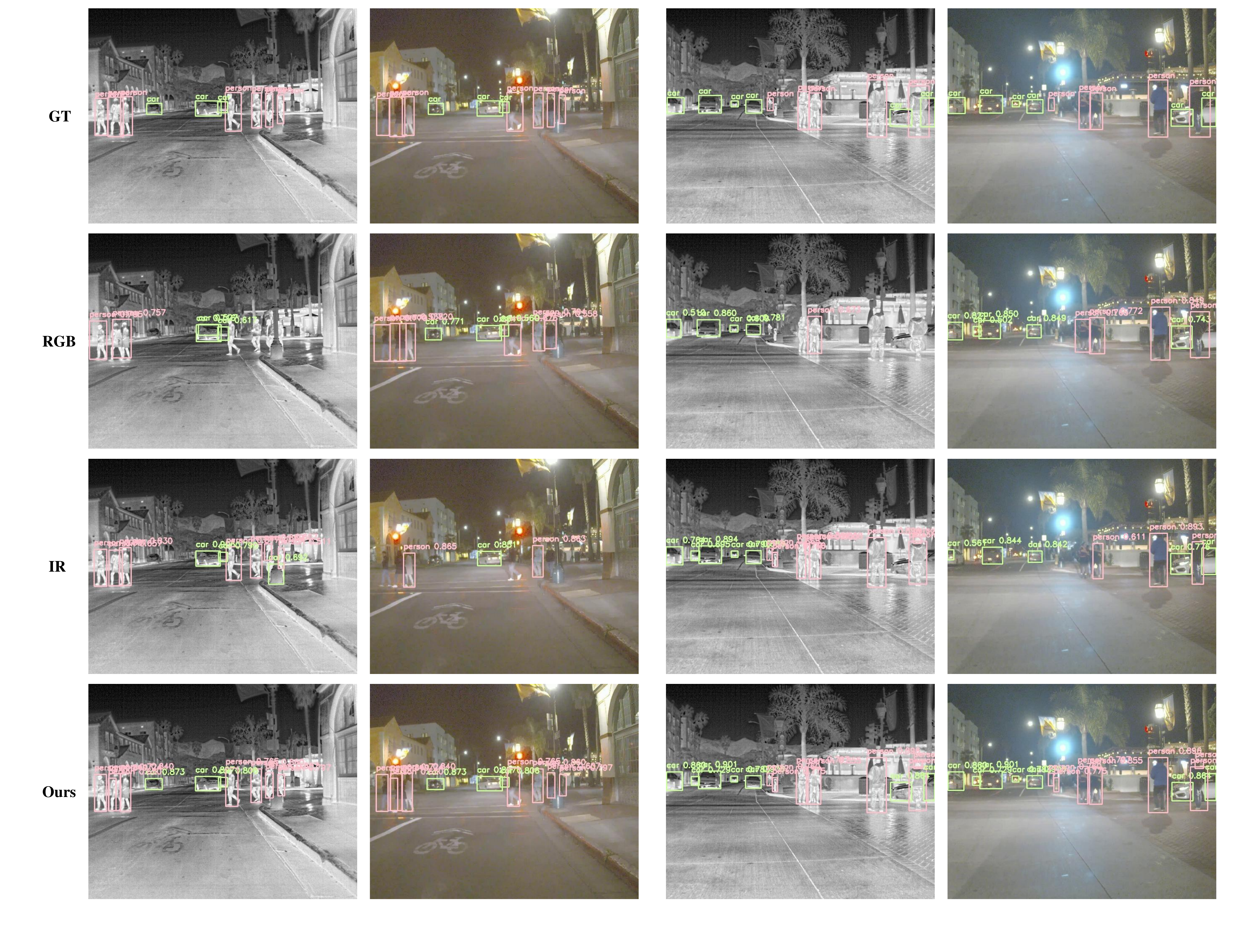}
    \caption{Qualitative results on FLIR using RGB, IR, and our fusion model.}
    \label{fig:visual_comparison}
        \vspace{-0.2cm}
 \end{figure}

\begin{table}[t]
\centering
\caption{Ablation study on the impact of backbone configurations for RGB and IR modalities on the FLIR-aligned dataset.}
\setlength{\tabcolsep}{2.8pt}
\begin{tabular}{c c | c c c | c c}
\toprule
\textbf{RGB} & \textbf{IR} & \textbf{mAP50} & \textbf{mAP75} & \textbf{mAP} & \textbf{Params (M)} & \textbf{MACs (G)} \\
\midrule
ResNet50 & ResNet18 & 79.2 & 40.0 & 43.6 & 56.04 & 87.97 \\
\rowcolor{gray!20}
ResNet18 & ResNet50 & \textbf{83.9} & \textbf{42.6} & 45.8 & 56.04 & 87.97 \\ 
ResNet50 & ResNet50 & 83.4 & 42.4 & \textbf{46.2} & 67.70 & 105.82 \\
\bottomrule
\end{tabular}
\label{tab:flir_ablation_backbone}
\vspace{-0.25cm}
\end{table}

\begin{table}[ht]
\centering
\caption{Ablation study on the contribution of each proposed component. Results are reported on the FLIR-aligned dataset.}
\renewcommand{\arraystretch}{0.9}
\begin{tabular}{ccc|cccc}
\toprule
\textbf{MSFD} & \textbf{CCSG} & \textbf{CLKG} & \textbf{mAP50} & \textbf{mAP75} & \textbf{mAP} \\
\midrule
& & & 80.7 & 40.2 & 44.1 \\
\checkmark & & & 81.0 & 41.3 & 44.7 \\
\checkmark & \checkmark & & 82.4 & 42.3 &  45.4 \\
\rowcolor{gray!20}
\checkmark & \checkmark & \checkmark & \textbf{83.9} & \textbf{42.6} & \textbf{45.8} \\
\bottomrule
\end{tabular}
\label{tab:ablation_performance}
\vspace{-0.25cm}
\end{table}

\subsection{Experimental Results and Analysis}

As shown in Table~\ref{tab:combined_flir_llvip}, our method achieves competitive performance on FLIR-aligned and achieves a state-of-the-art performance on LLVIP. Despite using a lightweight RGB backbone, IC-Fusion consistently outperforms prior fusion methods, demonstrating the effectiveness of our approach in challenging visual conditions. To understand the contribution of each modality, we perform an ablation study with different RGB-IR backbone configurations. As presented in Table~\ref{tab:flir_ablation_backbone}, using a stronger backbone for the IR modality (ResNet-50) and a lightweight one for RGB (ResNet-18) yields the best performance-to-complexity trade-off. In contrast, assigning a stronger backbone to the RGB modality leads to a notable performance drop, highlighting the importance of focusing on IR features. Interestingly, our asymmetric setup even outperforms the dual ResNet-50 configuration despite having fewer parameters and reduced computational cost. This result further validates the effectiveness of our IR-centric design. 

Table~\ref{tab:ablation_performance} presents a detailed ablation study on the contribution of each proposed component. The results clearly show that each module incrementally improves performance, validating their individual effectiveness and necessity within our fusion framework. We further analyze model complexity in Table~\ref{tab:params_macs_comparison}, comparing IC-Fusion against RT-DETR baselines. Multiply-accumulate operations (MACs) are computed using the \texttt{fvcore}~\cite{fvcore2023} library. Qualitative results on the FLIR dataset are visualized in Fig.~\ref{fig:visual_comparison}. Compared to unimodal RGB and IR detectors, IC-Fusion produces more accurate and stable predictions in low-light scenes, especially for small or occluded objects. This demonstrates the model's ability to effectively leverage modality complementarity for robust detection under real-world conditions.
\vspace{-0.15cm}


\begin{table}[t]
\centering
\caption{Comparison of Parameters and MACs with baseline.}
\renewcommand{\arraystretch}{0.9}
\begin{tabular}{l | c | c c}
\toprule
\textbf{Methods} & \textbf{Modality} & \textbf{Params (M)} & \textbf{MACs (G)} \\
\midrule
RT-DETR~\cite{zhao2024detrs} & Single & 42.70 & 68.81 \\
RT-DETR~\cite{zhao2024detrs} & Multi & 66.31 & 104.13 \\
\midrule
\rowcolor{gray!20}
IC-Fusion (ours) & Multi & 56.04 & 87.97 \\
\bottomrule
\end{tabular}
\label{tab:params_macs_comparison}
\vspace{-0.3cm}
\end{table}

\section{Conclusion}
In this letter, we proposed IC-Fusion, a multispectral object detection framework that leverages IR-centric feature fusion. To this end, we designed a lightweight yet effective fusion module that integrates complementary cross-modal features. Specifically, the module incorporates a Multi-Scale Feature Distillation (MSFD) block to enhance RGB features, followed by a cascade of a Cross-Modal Channel Shuffle Gate (CCSG), a Cross-Modal Large Kernel Gate (CLKG), and a Channel Shuffle Projection (CSP) module, which emphasize informative features from each modality while suppressing redundant information through gated cross-modal interaction. Extensive experiments on multispectral detection benchmarks validate the effectiveness of our design.

\bibliographystyle{IEEETran}
\bibliography{bibliography}

\end{document}